\documentclass[sigchi]{acmart}
\usepackage[font=scriptsize]{caption}
\usepackage{subcaption}
\usepackage{multirow}
\usepackage{color,soul}
\usepackage{enumitem}
\usepackage{booktabs}
\usepackage{ulem}
\AtBeginDocument{%
  \providecommand\BibTeX{{%
    \normalfont B\kern-0.5em{\scshape i\kern-0.25em b}\kern-0.8em\TeX}}}

\makeatletter
\def\subsubsection{\@startsection{subsubsection}{3}{0pt}%
                                 {-.5\baselineskip \@plus -2\p@ \@minus -.2\p@}%
                 {3.5\p@}{\@subsubsecfont}}
\makeatother

\copyrightyear{2019}
\acmYear{2019}
\acmConference[ICMI '19]{2019 International Conference on Multimodal Interaction}{October 14--18, 2019}{Suzhou, China}
\acmBooktitle{2019 International Conference on Multimodal Interaction (ICMI '19), October 14--18, 2019, Suzhou, China}
\acmPrice{15.00}
\acmDOI{10.1145/3340555.3353731}
\acmISBN{978-1-4503-6860-5/19/10}

\renewcommand\footnotetextcopyrightpermission[1]{} 
\begin{document}

%
\title[Confounders and Multimodal Emotion Classification]{Controlling for Confounders in Multimodal Emotion Classification via Adversarial Learning}
%
\author{Mimansa Jaiswal}
\affiliation{%
  \institution{University of Michigan}
}
\email{mimansa@umich.com}

\author{Zakaria Aldeneh}
\affiliation{%
  \institution{University of Michigan}
}
\email{aldeneh@umich.com}

\author{Emily Mower Provost}
\affiliation{%
  \institution{University of Michigan}
}
\email{emilykmp@umich.com}

%

%
\begin{abstract}
Various psychological factors affect how individuals express emotions. Yet, when we collect data intended for use in building emotion recognition systems, we often try to do so by creating paradigms that are designed just with a focus on eliciting emotional behavior. 
Algorithms trained with these types of data are unlikely to function outside of controlled environments because our emotions naturally change as a function of these other factors. 
In this work, we study how the multimodal expressions of emotion change when an individual is under varying levels of stress. We hypothesize that stress produces modulations that can hide the true underlying emotions of individuals and that we can make emotion recognition algorithms more generalizable by controlling for variations in stress. To this end, we use adversarial networks to decorrelate stress modulations from emotion representations.  We study how stress alters acoustic and lexical emotional predictions,  paying special attention to how modulations due to stress affect the transferability of learned emotion recognition models across domains. 
Our results show that stress is indeed encoded in trained emotion classifiers and that this encoding varies across levels of emotions and across the lexical and acoustic modalities. 
Our results also show that emotion recognition models that control for stress during training have better generalizability when applied to new domains, compared to models that do not control for stress during training. We conclude that is is necessary to consider the effect of extraneous psychological factors when building and testing emotion recognition models.
\end{abstract}
\vspace{-0.5em}
%
%

\vspace{-0.5em}
%

%
\maketitle
\section{Introduction}
Many extraneous psychological factors influence how individuals express and perceive emotions~\cite{paulmann2016psychological}.
However, most emotion recognition algorithms, rely on data collected in controlled laboratory environments (e.g.,~\cite{busso2008iemocap, busso2017msp}) where influences from such factors are either not present, or kept constant.
The performance of emotion recognition algorithms is likely to vary when applied to data where these external psychological factors are present.
In this work, we study how an extraneous psychological factor, stress, affects multimodal (acoustic+lexical) emotion classifiers.
Stress 
can affect how individuals produce and perceive emotion~\cite{paulmann2016psychological}. 
Yet, the effect of stress levels on the performance of state-of-the-art emotion recognition systems has not been explored.

Extraneous psychological factors can act as confounding factors, 
variables that influence both the output (e.g., emotion) and the input (e.g., acoustic and lexical features).
Not controlling for confounding variables when training emotion classifiers
can cause the classifiers to learn unintentional associations
between the variables, associations that might not replicate in real world scenarios.
For instance, consider a dataset where all the ``sad'' samples were unintentionally recorded from individuals who were experiencing stress at the time of recording.
Not taking special care when building the models could cause a trained classifier to erroneously associate experiencing stress with being sad. 
In this work, we study how stress alters the performance of trained emotional classifiers in the context of neural networks. We then see how performance is affected when tested on samples out of domain, when we explicitly impede the network from learning such associations.

Previous research showed that controlling for confounding variables when training emotion recognition classifiers results in more robust models when compared to models trained without controlling for the same confounding variables. For instance, Abdelwahab et al.~\cite{abdelwahab2018domain} and Gideon et al.~\cite{gideon2019barking} showed that controlling for domain (i.e., data source), as a confounding factor, when training emotion recognition models results in improved cross-corpus generalization performance when compared to performance of models that were trained without controlling for domain as a confounding factor. 
Most of the above mentioned methods rely on samples obtained from the target domain to extract representations that retain information only about emotion and not domains. Our goal is to go beyond studying the effects of variations due to domain and background noise on the robustness of trained emotion recognition models, and instead
focus on how stress affects the learned acoustic and lexical emotional representations. Unlike the commonly used methods for learning domain invariance, we aim to accomplish generalizing person specific behavior by proactively ``unlearning'' the modulations due to the presence of stress while still retaining emotion information in representations.

In particular, we seek to answer the following questions:
\begin{enumerate}
\vspace{-0.3em}
\item Can we recognize stress given representations trained solely for recognizing emotion? Is the stress recognition performance similar across the lexical and acoustic modalities?
\item Can we completely decorrelate emotion representations from stress representations? If so, how does this decorrelation impact the performance of emotion classifiers? 
\item Does the impact of decorrelation on the performance of emotion classifiers vary given different levels of stress?
\item Does decorrelating these representations (i.e., emotion and stress) aid in model generalizability?
\item Can we proactively remove other types of confounders (e.g., spontaneity) to improve cross-dataset performance?
\item Are there identifiable lexical patterns in samples that are especially successfully classified by the adversarially trained model for emotion classification?
\vspace{-0.3em}
\end{enumerate}
To the best of our knowledge, this is the first work that studies the interplay between emotion and stress in the context of automatic emotion recognition and representation learning.

\vspace{-0.75em}
\section{Related Work}
Previous research has looked at removing confounding factors as a graph problem, with methods such as graph pruning~\cite{molchanov2016pruning}, surgery estimation~\cite{subbaswamy2018learning} and counterfactual adjustments. These methods are usually limited to data in low feature space unlike speech or language representations. In the context of neural networks, controlling for confounding factors during training is commonly achieved via the adversarial training paradigm
\cite{ganin2016domain, mchardy2019adversarial, elazar2018adversarial, moore2018modeling, meng2018speaker, zhang2018mitigating}.
In this paradigm, a network learns how to perform a specific task (e.g., detect emotion) while at the same time ``unlearns'' how to perform another task (e.g., detect stress).

One group of methods have considered confounding factors that are either singularly labeled or cannot be labeled. 
Ben-David et. al~\cite{ben2010theory} showed that a classifier trained to predict the sentiment of reviews can implicitly learn to predict the category of the products. The authors used an adversarial multi-task classifier to learn domain invariant sentiment representations.
Shinohara~\cite{shinohara2016adversarial} used an adversarial approach to train noise-robust networks for automatic speech recognition.
They used domain (i.e., background noise) as the adversarial task while training the model to obtain representations that are both senone-discriminative and domain-invariant.
In emotion recognition applications, Abdelwahab et al.~\cite{abdelwahab2018domain} used domain adversarial networks to improve cross-corpus generalization for emotion recognition tasks.

Another group of methods handles confounding factors that were explicitly labeled during the data collection process.
Meng et al.~\cite{meng2018speaker} used adversarial multi-task learning to curtail variances due to speaker identity when training automatic speech recognition systems, demonstrating how controlling for such variations improves generalization performance.
McHardy et al.~\cite{satire-adv} used the same approach to prevent the network from learning publication source characteristics while being primarily trained for recognizing instances of satire. They demonstrated how classifiers trained to predict satire often predict publication source, by associating the publication source to the intended target label.

Adversarial multi-task learning was also used in other fields (e.g., computer vision, language processing) to train models that are invariant to certain properties, yet discriminative with respect to others~\cite{ganin2016domain, mchardy2019adversarial, elazar2018adversarial}.
Given a source domain and a target domain, the goal is to study how controlling for an extraneous confounding factor 
when training our emotion recognition models on the source domain affects the performance of the trained models on both source and target domains. Previous research has shown that although controlling for extraneous confounders while training models causes the performance to drop on the source domain, it improves the performance on the target domain~\cite{mchardy2019adversarial, tzeng2017adversarial}.

\vspace{-7pt}
\section{Datasets and Features}
\subsection{Datasets}
We use three datasets to study the effect of stress on emotion recognition:
(1) Multimodal Stressed Emotion (MuSE) dataset~\cite{jaiswal2019muse};
(2) Interactive Emotional Dyadic MOtion Capture (IEMOCAP) dataset~\cite{busso2008iemocap}; 
and (3) MSP-Improv dataset~\cite{busso2017msp}.

\textbf{MuSE.} The MuSE dataset was collected to understand the interplay between stress and emotion in natural spoken communication. 
The dataset consists of 55 recordings from 28 participants.
Each participant in the dataset was recorded across two sessions: stressed and not-stressed (one person only participated in the stressed condition).
Stress was elicited by recording data during the final exam period at the 
University of Michigan 
while data for the not-stressed condition were recorded after exams concluded.
Emotions were induced in the participants via video stimuli and 
via emotionally evocative monologue topics~\cite{aron1997experimental}.
Utterances were created in the dataset by identifying prosodic or lexical boundaries in spontaneous speech in the monologues as defined in~\cite{kolavr2008automatic}.
The final dataset that we use in this study has a total duration of around 10 hours from 2,648 utterances. Data selection was performed to exclude utterances less than three seconds and greater than 35 seconds. The assumption behind this exclusion criteria was that shorter utterances may not capture enough emotional information while longer utterances may capture varying emotional information~\cite{khorram2018priori}.

\textbf{IEMOCAP.} The IEMOCAP dataset was created to explore the relationship between emotion, gestures, and speech. Pairs of actors, one male and one female (five males and five females in total), were recorded over five sessions. Each session consisted of a pair performing either a series of given scripts or improvisational scenarios. 
The data were segmented by speaker turn, resulting in a total of 10,039 utterances (5,255 scripted turns and 4,784 improvised turns). 

\textbf{MSP-Improv.} The MSP-Improv dataset was collected to capture naturalistic emotions from improvised scenarios. It partially controlled for lexical content by including target sentences with fixed lexical content that are embedded in different emotional scenarios.
The data were divided into 652 target sentences, 4,381 improvised turns (the remainder of the improvised scenario, excluding the target sentence), 2,785 natural interactions (interactions between the actors in between recordings of the scenarios), and 620 read sentences 
for a total of 8,438 utterances.

\vspace{-1em}
\subsection{Labels}

\textbf{Emotion Labels.}  Each utterance in the MuSE dataset was labeled for  \textit{activation} 
and \textit{valence} on a nine-point Likert scale by eight crowd-sourced 
annotators~\cite{jaiswal2019muse}, who observed the data in random order across subjects. We average the annotations to obtain a mean score for 
each utterance, and then bin the mean score into one of three classes, defined as, \{``\textit{low}'': [min, 4.5], 
``\textit{mid}'': (4.5, 5.5], ``\textit{high}'': (5.5, max]\}. 
The resulting distribution for activation is: \{``\textit{high}'': $24.58\%$, ``\textit{mid}'': $40.97\%$ and ``\textit{low}'': $34.45\%$\} and for valence is \{``\textit{high}'': $29.16\%$,  ``\textit{mid}'': $40.44\%$ and ``\textit{low}'': $30.40\%$\}.
Utterances in IEMOCAP and MSP-Improv were annotated for valence and activation on a five-point Likert scale.
The annotated activation and valence values were averaged for an utterance and binned as: \{``\textit{low}'': [1, 2.75], ``\textit{mid}'': (2.75, 3.25], ``\textit{high}'': (3.25, max]\}

\textbf{Stress Labels.} Utterances in the the MuSE dataset include stress annotations, in addition to the activation and valence annotations.
The stress annotations for each session were self-reported by the participants using the Perceived Stress Scale (PSS)~\cite{cohen1983global}. 
We perform a paired t-test for subject wise PSS scores, and find that the scores are significantly different for both sets (16.11 vs 18.53) at $p<0.05$. This especially true for question three 
(3.15 vs 3.72), and hence, we double the weightage of the score for this question while obtaining the final sum.
We bin the original nine-point adjusted stress scores into three classes, \{``\textit{low}'': (min, mean$-2$], ``\textit{mid}'': (mean$-2$, mean$+2$], ``\textit{high}'': (mean$+2$, max]\}.
We assign the same stress label to all utterances from the same session. The distribution of our data for stress is {``\textit{high}'': $40.33\%$, ``\textit{mid}'': $25.78\%$ and ``\textit{low}'': $38.89\%$}

\textbf{Improvisation Labels.} Utterances in the IEMOCAP dataset were recorded in either a scripted scenario or an improvised one. We label each utterance with a binary value \{``\textit{scripted}'', ``\textit{improvised}''\} to reflect this information.

\vspace{-1.0em}
\subsection{Features}
The goal is to study the effect of stress on trained multimodal (acoustic and lexical) emotion classifiers.

\textbf{Acoustic.}  We use Mel Filterbank (MFB) features, which are frequently used in speech  processing applications, including speech recognition, 
and emotion recognition~\cite{khorram2017capturing, krishna2018study}.  We extract the 40-dimensional MFB features using a 25-millisecond Hamming window with a step-size of 10-milliseconds. As a result, each utterance is represented as a sequence of 40-dimensional feature vectors. We $z$-normalize the acoustic features by session for each speaker.

\textbf{Lexical.} We have human transcribed data available for MuSE and IEMOCAP. We use the word2vec representation based on these transcriptions, which has shown success in sentiment and emotion analysis tasks~\cite{kim2014convolutional}. 
We represent each word in the text input as a 300-dimensional vector 
using a pre-trained word2vec model~\cite{mikolov2013distributed}, replacing out-of-vocab
words with the $\langle unk\rangle$ token.
Each utterance is represented as a sequence of 300-dimensional feature vectors. We use just acoustic inputs for MSP-Improv because human transcriptions are not available.

\section{Experimental Setup}
In this section, we describe the network architecture and the training recipe of the two emotion recognition models, one that controls for 
stress as a confound and one that does not.

\begin{figure}[t]
  \centering
  \includegraphics[width=0.8\linewidth]{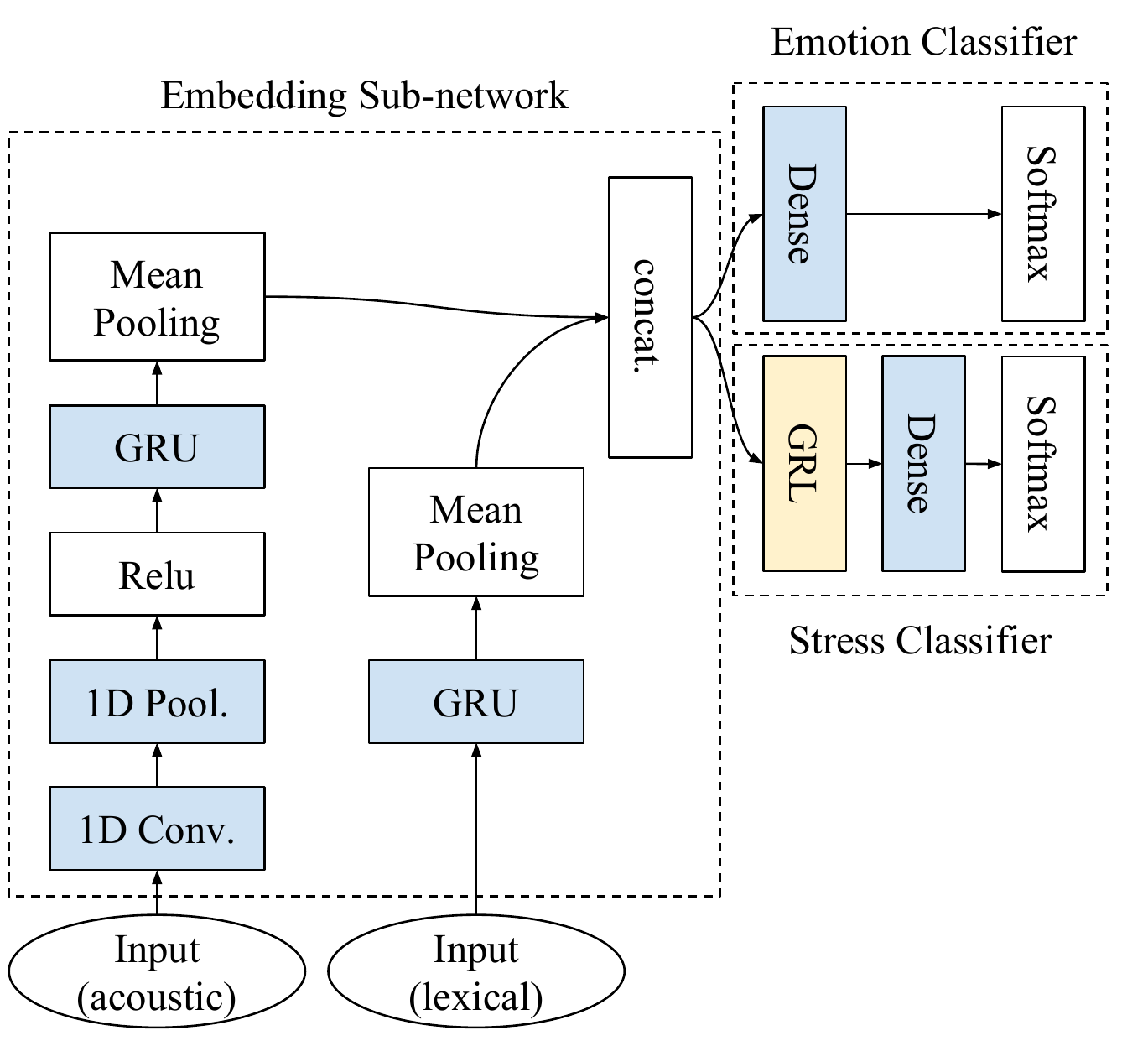}
  \caption{Adversarial multi-task network architecture.}
  \label{fig:model}
  \vspace{-10pt}
\end{figure}
\vspace{-1.0em}
\subsection{Architecture}
\label{sec:setup_arch}

The network consists of three components (Figure~\ref{fig:model}): (1) embedding sub-network; 
(2) emotion classifier; and (3) stress classifier.
The embedding sub-network induces fixed-size representations given the 
acoustic and lexical input streams. In Figure~\ref{fig:model}, the concatenation layer of acoustic and lexical stream shows the induced fixed-size representations. The emotion and stress classifiers
perform their respective classification tasks given the fixed-size representations from
the embedding sub-network. 
We use two variants of the embedding sub-network in this work: a unimodal
and a multimodal variant.
The unimodal embedding sub-network takes a single stream (acoustic or lexical) input while the multimodal embedding sub-network takes a two-stream (acoustic and lexical)
input.
The objective of the adversarial multi-task system is to maximize the performance of the emotion classifier while minimizing the performance of the stress classifier.  

\textbf{Stress-Invariance.}
The network is trained to unlearn stress.  We achieve this goal using a Gradient Reversal Layer (GRL)~\cite{ganin2014unsupervised}.
The use of GRLs is a common approach that can be used to train networks that are invariant to specific properties
\cite{meng2018speaker, shinohara2016adversarial, abdelwahab2018domain, ganin2016domain, mchardy2019adversarial, elazar2018adversarial}.
During the backward pass of the training phase, the GRL
multiplies the backpropagated gradients by $-\lambda$.
During the forward pass, the GRL acts as an identity function.
To make the network invariant to stress, we place the GRL between 
the embedding sub-network and the stress classifier as shown in 
Figure~\ref{fig:model}.

\textbf{Model Variations.} We use 12 variants of the network shown in Figure~\ref{fig:model} with the following combinations:
\textit{\{normal-classification, adversarial-classification\}} $\times$ \textit{\{activation, valence\}} $\times$
\textit{\{uni-lexical, uni-acoustic, multimodal\}}.
The normal classification setup consists of the embedding sub-network (lexical, acoustic, or multimodal) and the emotion classifier (activation or valence) parts of the model.
The adversarial classification setup adds the adversarial stress classifier.
\subsection{Training}
We implement models using the Keras library~\cite{keras2015}.
We use a weighted cross-entropy loss function for each task and learn the model parameters using the RMSProp optimizer~\cite{rmsprop}.
We train our networks for a maximum of 50 epochs and monitor the validation loss from the emotion classifier after each epoch, stopping the training if the validation loss does not improve after five consecutive epochs.
Once the training process ends, we revert the network's weights to those that
achieved the lowest validation loss on the emotion classification task. 
For the adversarial classification model, we ensure that the chosen model yields a validation unweighted average recall (UAR) that is random (0.33) for the stress classification task. 
Finally, we train each setup three times with different random seeds and average the predictions over these runs to reduce variations due to random initialization.

We use validation samples for hyper-parameter selection and early stopping. 
The hyper-parameters that we use in our search include:
number of convolutional layers \{3, 4\}, 
number of convolutional kernels \{2, 3\}, 
conv. layers width  \{32, 64, 128\}, 
1D max-pooling kernel width  \{2\}, 
number of GRU layers  \{2, 3\}, 
GRU layers width  \{32\}, 
number of dense layers  \{1, 2\}, 
dense layers width  \{32, 64\}, 
GRL $\lambda$  \{0.3, 0.6, 0.8\}.
For the adversarial emotion classification setups, we use the hyper-parameters that maximize the validation emotion classification performance while minimizing the validation stress classification performance.
We assess performance using UAR, given the imbalanced nature of our data~\cite{rosenberg2012classifying}.

\section{Analysis}

\subsection{Question 1}
\label{sec:analysis_1}

\begin{table*}
\scriptsize
           \centering
           \captionsetup[subtable]{position = below}
          \captionsetup[table]{position=top}
          \label{exp-perf}
\caption{\footnotesize UAR (chance = 0.333) for predicting activation (left) and valence (right) in adversarial and non-adversarial (normal) setups. Bold signifies significantly different performance (paired $t$-test, $\alpha<0.05$).}
  
\begin{subtable}{\columnwidth}
\setlength{\tabcolsep}{0.5em}
              \centering
                     \begin{tabular}{lcccc}
     \toprule
     & \multicolumn{2}{c}{Normal} & \multicolumn{2}{c}{Adversarial}  \\
     \cmidrule(r){1-1} \cmidrule(r){2-3} \cmidrule(r){4-5}
          Setup & Act.     &  Stress  & Act.     &  Stress\\
     \midrule
     Unimodal (A) & \textbf{0.611} & 0.412 & \textbf{0.572}  & 0.305    \\
     Unimodal (L) & 0.550  & 0.394  & 0.527 & 0.332       \\
     Multimodal (A+L)   & \textbf{0.659}  & 0.425 & \textbf{0.613}  & 0.322      \\
     \bottomrule
   \end{tabular}
               \caption{\footnotesize Activation}
               \label{exp-performance-act}
           \end{subtable}%
           \hspace*{4em}
\begin{subtable}{\columnwidth}
\setlength{\tabcolsep}{0.5em}
               \centering
                  \begin{tabular}{lcccc}
                    \toprule
                    & \multicolumn{2}{c}{Normal} & \multicolumn{2}{c}{Adversarial}  \\
                    \cmidrule(r){1-1} \cmidrule(r){2-3} \cmidrule(r){4-5}
                        Setup & Val.     &  Stress  & Val.     &  Stress\\
                    \midrule
                    Unimodal (A) & 0.460 & 0.396 & 0.431  & 0.332    \\
                    Unimodal (L)   & 0.685  & 0.353 & 0.674  & 0.323      \\
                    Multimodal (A+L)   & \textbf{0.666}  & 0.397 & \textbf{0.641}  & 0.328      \\
                    \bottomrule
                  \end{tabular}    
                \caption{\footnotesize Valence}
                 \label{exp-performance-val}
           \end{subtable}
    \vspace{-6pt}       
       \end{table*}
%

\begin{table*}[t]
\scriptsize
\caption{\footnotesize Confusion matrices for activation (left) and valence (right) showing percentage change in classification performance of the multimodal setup for each emotion class after controlling for stress.}
\begin{subtable}{\columnwidth}\centering
\setlength{\tabcolsep}{0.65em}
\begin{tabular}{l|ccc|ccc}
    \toprule
    & \multicolumn{3}{c}{Low Stress} & \multicolumn{3}{c}{High Stress}  \\
    \cmidrule(r){2-4} \cmidrule(r){5-7}
    Act.     & (0) & (1) & (2) & (0) & (1) & (2) \\
    \midrule
    (0) & $-1.21$ & $-0.22$ & $+1.44$ & $+1.30$ & $-2.11$ & $+7.35$ \\
    (1)  & $+4.22$ & $-2.73$ & $+1.31$ & $+18.40$ & $-22.03$ & $+14.80$\\
    (2)  & $-2.01$ & $+6.66$ & $-6.38$ & $-8.81$ & $+6.21$ & $-3.14$\\
    
    \bottomrule
  \end{tabular}
  \caption{\footnotesize Activation}
\end{subtable}
\quad
\hfill
\begin{subtable}{\columnwidth}\centering
\setlength{\tabcolsep}{0.6em}
\begin{tabular}{l|ccc|ccc}
    \toprule
    & \multicolumn{3}{c}{Low Stress} & \multicolumn{3}{c}{High Stress}  \\
    \cmidrule(r){2-4} \cmidrule(r){5-7}
    Val.     & (0) & (1) & (2) & (0) & (1) & (2) \\
    \midrule
    (0) & $-1.66$ & $+1.12$ & $+1.11$ & $-8.22$ & $+7.88$ & $+1.33$ \\
    (1)  & $-0.76$ & $-2.22$ & $+0.87$ & $-2.31$ & $-6.11$ & $+4.26$ \\
    (2)  & $-1.01$ & $+0.31$ & $+1.45$ & $-1.15$ & $-2.10$ & $+0.79$\\
    
    \bottomrule
  \end{tabular}
  \caption{\footnotesize Valence}
\end{subtable}
\label{emotion-confusion-matrix}
\vspace{-6pt}
\end{table*}

\uline{Question}: \textit{Can we recognize stress given representations trained solely for recognizing emotion?}\\
\uline{Hypothesis}: \textit{We expect the performance of detecting stress from representations obtained by training emotion classifiers to vary depending on the modality, and the emotion being modeled.}\\
Stress has been shown to have varying effects on both the linguistic~\cite{buchanan2014acute} and para-linguistic~\cite{paulmann2016psychological, steeneken1999speech} components of
communication. Previous work has also demonstrated that the lexical part of speech carries more information about valence while the para-linguistic part carries more information about activation~\cite{khorram2017capturing}.
As a result, we expect the performance of stress 
classification to vary based on modality, and emotion dimension being modeled. 

To test our hypothesis, we train the 12 model variants described in section~\ref{sec:setup_arch} with five-fold speaker-independent cross-validation.
We report the average across the five folds for the normal classification and the adversarial classification setups in 
Tables~\ref{exp-performance-act} and~\ref{exp-performance-val} for predicting activation and valence, respectively. 
Our results show that a network trained to only recognize emotion is generally discriminative for stress. For instance, we obtain a maximum UAR of 0.425 when using a multimodal network that was trained to only detect activation; and a maximum UAR of 0.397 when using multimodal network that was trained to only detect valence.

Our results in Table~\ref{exp-performance-act} suggest that the acoustic modality encodes information that is relevant for recognizing stress and activation. 
In contrast, the results show that the representations trained on lexical modality encode information that is relevant for detecting valence but not for stress.
Our findings are consistent with previous research that demonstrated that stress is encoded in acoustic features \cite{buchanan2014acute, freeman2009comprehension, dedora2011acute}. 

\vspace{-6pt}
\subsection{Question 2}
\label{sec:analysis_2}

\uline{Question}: \textit{Can we decorrelate emotion representations from stress representations? How does it impact performance of emotion classifiers?}\\
\uline{Hypothesis}: \textit{Decorrelating the stress and emotion representations will cause a decrease in emotion classification performance on the source domain.}\\
Previous research demonstrated that controlling for confounders during the training process can cause the performance of the main task on the same dataset to decrease~\cite{satire-adv,subbaswamy2018learning}.
For instance, Zhang et al.~\cite{satire-adv} showed that the performance of detecting sarcasm decreases when controlling for publication as a confounding variable in the training process, but the prediction accuracy increases on an unseen publication set. Similarly, Ganin et al.~\cite{ganin2016domain} showed that controlling for domain while training a network for detecting sentiment can result in a performance reduction on the main task.
The reduction in performance on the main task, after controlling for an extraneous confounding variable, can be attributed to the removal of information that the model can use as a ``shortcut'' for achieving the main task.

Our results show that (Tables~\ref{exp-performance-act} and~\ref{exp-performance-val}):
\begin{itemize}
  \item Activation classification performance decreases given adversarial training.  This decrease is statistically significant for the acoustic (6.382\% drop in UAR) and multimodal (6.980\% drop in UAR) setups.
  \item Valence classification performance decreases given adversarial training. This decrease is statistically significant for the multimodal setup (3.754\% drop).
\end{itemize}
The reduction in performance in the main task after controlling for a confounding variable can also be caused by the removal of information that is equally beneficial for both detecting stress and detecting emotion.
Our results in 
further sections
, however, show that models that control for stress are better able to recognize emotion in new domains, compared to models that do not control for stress. This suggests that the process of ``unlearning'' stress does not come at the expense of the primary task of emotion detection.

\vspace{-6pt}
\subsection{Question 3}
\label{sec:analysis_3}
\uline{Question}: \textit{Does the impact on the performance of emotion classifiers vary given different levels of stress}\\
\uline{Hypothesis}: \textit{The valence and activation emotion classes (low, medium, and high) are impacted differently by stress.}\\
Prior research demonstrated that emotions produced by stressed individuals are not recognized in the same way as 
those
by non-stressed individuals~\cite{paulmann2016psychological}. In particular, researchers  found that speech patterns of negative emotions produced by stressed individuals are more difficult to recognize than negative emotions produced by non-stressed individuals \cite{paulmann2016psychological}. 
We expect similar patterns to hold in automatic emotion recognition systems.
That is, we expect the presence of stress to have a varying effect on the performance of the classifier depending on the emotion class (for valence and activation), and the amount of stress induced.

To test our hypothesis, we study how the performance of the classifier varies for each emotion class when we control for
stress. We report the changes in performance, after controlling for stress, for each emotion class, grouped by stress level
(low, high), in Table~\ref{emotion-confusion-matrix}. The results in the table show:

\begin{itemize}
\vspace{-0.3em}
\item High levels of stress impact classification more strongly (3.89\% and 2.41\% drop in UAR for activation and valence, respectively) than low levels of stress do (1.44\% and 0.31\% drop in UAR for activation and valence, respectively). This is generally true for all emotion classes (valence and activation).
\item High levels of stress have the biggest impact on mid level of activation predictions (22.03\% drop in accuracy for detecting neutral activation).
\item High levels of stress have the biggest impact on low valence predictions (8.22\% drop for low valence).

\end{itemize}
\vspace{-0.3em}
The results show that stress level effects emotion recognition, for both activation and valence. 
This drop in performance can be attributed to changes in the perceived emotions by the annotators. 
Researchers 
have
demonstrated that stressed sentences are usually perceived by annotators to be more neutral than they were originally intended to be \cite{paulmann2016psychological}. 


\vspace{-6pt}
\subsection{Question 4}
\label{sec:analysis_4}
\uline{Question}: \textit{Does the process of decorrelating these representations (i.e., emotion and stress) aid in model generalizability?}\\
\uline{Hypothesis}: \textit{Removing the confounding factor stress would aid in creating models that are more generalizable across datasets.}\\ Previous research has shown that laboratory collected datasets are too small and often fail to capture the complete distribution of the domain~\cite{hu2017frankenstein, locatello2018challenging} present in the real world. These datasets are often plagued with unintentional correlational factors~\cite{locatello2018challenging, landeiro2016robust}. Hence we believe that removing modulations due to stress should aid the generalizability of the model to datasets, where this psychological factor is either unmeasured or the distribution is non-uniform between training and testing.

To answer if the models trained on MuSE dataset generalize better, we perform two sets of experiments: (a) self generalizability in artificially partitioned datasets with different stress distributions for evidence of concept and (b) cross-dataset generalizability.

\textbf{Artificially Segmented Within-Dataset Performance.} We run the first set of experiments by creating partitions of data by stress level. We do this to create artificially mismatched environments between training and testing. 
We reserve one set to be test set (target), while keeping the other two for training and validation combined (source). This is in similar vein to partitioning created across confounding factor for UCI Bike Rentals Dataset in \cite{subbaswamy2018learning}. To ensure speaker independent sets, we divide the training set using an 80:20  split (train and validation), ensuring no speakers overlap. We run these models $n$ times where $n$ is the number of speakers in test data, that are also present in train/validation data. For each run, we remove one speaker from the train/validation data and test on that speaker. We calculate the average test performance of all these runs as the performance of the model for that setup.

\begin{table}[]
\scriptsize
\caption{\footnotesize Performance (UAR)  predicting activation (left) and valence (right) in non-adversarial and adversarial (for best \textit{lambda} value) for self-partition on MuSE. Bold signifies significantly different performance (paired $t$-test, $\alpha<0.05$).
  }
  \label{exp-tri-self-partition}
\centering
\begin{tabular}{lllll}
\toprule
           & \multicolumn{2}{c}{Normal} & \multicolumn{2}{c}{Adversarial} \\ \cmidrule(r){2-3} \cmidrule(r){4-5}
                 & Act            & Val            & Act              & Val             \\ \midrule
\multicolumn{5}{c}{\textbf{Train: Stress (Medium + High) Test: Stress (Low)}}                                                      \\ \midrule
Unimodal (A)     &       \textbf{0.623}        &       0.451         &        0.582          &     0.453            \\
Unimodal (L)     &      0.561          &      0.654         &       0.548           &       \textbf{0.691}          \\
Multimodal (A+L) &       0.650         &        0.673        &        \textbf{0.662}          &       \textbf{0.703}          \\ \midrule
\multicolumn{5}{c}{\textbf{Train: Stress (Low + High) Test: Stress (Medium)}}                                                      \\ \midrule
Unimodal (A)     &      0.610          &        0.420        &       \textbf{0.628}           &    \textbf{0.432}             \\
Unimodal (L)     &        0.520        &      0.672         &      \textbf{0.545}            &      0.669           \\
Multimodal (A+L) &      0.602          &        0.621        &        \textbf{0.638}          &   \textbf{0.649}              \\ \midrule
\multicolumn{5}{c}{\textbf{Train: Stress (Medium + High) Test: Stress (Low)}}                                                      \\ \midrule
Unimodal (A)     &      0.582          &     0.384           &        \textbf{0.605}          &   \textbf{0.411}              \\
Unimodal (L)     &    0.540            &     0.630          &       0.513           &      \textbf{0.652}           \\
Multimodal (A+L) &        0.642        &     0.621           &      0.647            &     \textbf{0.637}            \\ 
 \bottomrule
\end{tabular}
\vspace{-2em}
\end{table}

We report our results in Table~\ref{exp-tri-self-partition}.
When we consider low levels of stress as our target, we see that adversarial classification significantly improves performance over normal classification for multimodal setup for activation and for both, lexical and multimodal setup for valence.
Considering mid levels of stress as our target, adversarial classification significantly improves performance over normal classification for all setups for activation and for both, acoustic and multimodal setup for valence.
Subsequently considering high levels of stress as our target, adversarial classification significantly improves performance over normal classification for acoustic setups for activation and for all setups for valence.

\textbf{Cross-dataset Performance.} Now that we have evidence for concept for artificially mismatched distributions that removing stress as a confounding factor can aid generalizability, we ask if adversarially removing encoded stress from emotion representation improves cross-dataset performance. We assume that we previously do not have any samples from the target dataset to train our model, to test generalizability at deployment. We train a dataset on complete MuSE data, keeping 20\% of speaker independent data for validation of hyper-parameters. Then we use the trained model to test on IEMOCAP for a combination of acoustic, lexical and multimodal inputs and on MSP-Improv for acoustic inputs.

We report our results of comparing the performance of the adversarial and normal models (Table~\ref{exp-tri-other}):
\begin{itemize}
    \item Activation: There is a significant increase in performance in all setups when the adversarial classification model is tested on IEMOCAP.
    We observe a significant increase in performance in acoustic setup of adversarial classification model (0.404 vs 0.421) when tested on MSP-Improv.
    \item Valence: There is a significant increase in performance using acoustic setup (0.376 vs 0.401) and multimodal setup (0.431 vs 0.472) of adversarial model when tested on IEMOCAP.
    We see no significant difference in performance when testing on MSP-Improv using adversarial classification model .
\end{itemize}
Based on these results, we understand that removal of a psychological confounding factor, stress, generally aids in the generalizability of the model on completely unseen data, where the distribution of this confounding factor is unknown. 

\begin{table}[]
\scriptsize
\caption{\footnotesize Performance (in UAR)  predicting activation (left) and valence (right) in  non-adversarial and adversarial (for best \textit{lambda} value) setups across datasets when trained on MuSE. Bold signifies significantly different performance (paired $t$-test, $\alpha<0.05$).
  }
  \label{exp-tri-other}
\centering
\begin{tabular}{lllll}
\toprule
           & \multicolumn{2}{c}{Normal} & \multicolumn{2}{c}{Adversarial} \\ \cmidrule(r){2-3} \cmidrule(r){4-5}
                 & Act            & Val            & Act              & Val             \\ \midrule
\multicolumn{5}{c}{\textbf{MuSE to IEMOCAP}}                                                      \\ \midrule
Unimodal (A)     &      0.419          &      0.376          &   \textbf{0.448}               &       \textbf{0.401}          \\
Unimodal (L)     &    0.401            &    0.433       &   \textbf{0.436}               &           0.447      \\
Multimodal (A+L) &      0.422          &      0.431          &      \textbf{0.459}            &   \textbf{0.472}              \\ \midrule
\multicolumn{5}{c}{\textbf{MuSE to MSP-Improv}}                                                   \\ \midrule
Unimodal (A)     &      0.404          &      0.368          &          \textbf{0.431}        &   0.372              \\ \bottomrule
\end{tabular}
\vspace{-1em}
\end{table}

\vspace{-6pt}
\subsection{Question 5}
\label{sec:analysis_5}

\uline{Question}: \textit{Can we proactively remove other types of confounders to improve cross-dataset performance?}\\
\uline{Hypothesis}: \textit{Removing the confounding factor of spontaneity in IEMOCAP will improve cross-dataset performance.}\\
We observed in 
the last question
that ``unlearning" the confounder stress can aid generalizability. Now, we want to see if the same method can be used to make models trained using other datasets more reliable to change in target data distribution. 
We hypothesize that decorrelating the effect of spontaneity on emotion representation will lead to models that generalize better. This is because, as shown in \cite{mangalam2017learning}, the emotional content expression is different in scripted vs spontaneous speech, and hence should modulate the emotion representation in trained model. To this end, we use the IEMOCAP dataset which has utterances that come from sessions that are both scripted and improvised. We do not use MSP-Improv for similar analysis here, because the scripted sessions, by corpus design, have limited lexical content and hence wouldn't cover enough input representation space for generalizability. We train the same 12 model variants described in section~\ref{sec:setup_arch} replacing the adversarial stress classifier sub-component with adversarial spontaneity classifier for this analysis.

We report our results in Table~\ref{exp-iemocap-other}. We compare the performance of the adversarial and normal models:
\vspace{-0.1em}
\begin{itemize}
    \item Activation: There is a significant increase in performance in lexical (0.401 vs 0.425) and multimodal setup (0.433 vs 0.467) when the adversarial model is tested on MuSE dataset.
    We see no significant difference in performance when the adversarially trained model is tested on MSP-Improv.
    \item Valence: There is a significant increase in the performance using all setups of the adversarial classification model when tested on MuSE.
    We observe a significant increase in the performance in the acoustic setup of the adversarial model (0.410 vs 0.438) when tested on MSP-Improv.
\end{itemize}
\vspace{-0.1em}
We see that the removal of modulations due to the data collection methodology improves generalizability for many cross-dataset cases. This suggests that this method can be extended to train stabler models by explicitly accounting for confounding variables in limited amounts of training data.

\begin{table}[t]
\scriptsize
\caption{\footnotesize Performance (in UAR)  predicting activation (left) and valence (right) in non-adversarial and adversarial (for best \textit{lambda} value) setups across datasets when trained on IEMOCAP. Bold signifies significantly different performance (paired $t$-test, $\alpha<0.05$).
  }
 \label{exp-iemocap-other}
\centering
\begin{tabular}{lllll}
\toprule
           & \multicolumn{2}{c}{Normal} & \multicolumn{2}{c}{Adversarial} \\ \cmidrule(r){2-3} \cmidrule(r){4-5}
                 & Act            & Val            & Act              & Val             \\ \midrule
\multicolumn{5}{c}{\textbf{IEMOCAP to MuSE}}                                                      \\ \midrule
Unimodal (A)     &       0.428         &      0.401          &   0.427               &      \textbf{0.431}           \\
Unimodal (L)     &      0.401          &    0.423            &     \textbf{0.425}             &         \textbf{0.455}        \\
Multimodal (A+L) &     0.430           &       0.429         &        \textbf{0.463}          &          \textbf{0.468}       \\ \midrule
\multicolumn{5}{c}{\textbf{IEMOCAP to MSP-Improv}}                                                   \\ \midrule
Unimodal (A)     &       \textbf{0.414}         &      0.410          &       0.402           & \textbf{0.439}           \\ \bottomrule
\end{tabular}
\vspace{-1em}
\end{table}

\vspace{-6pt}
\subsection{Question 6}
\label{sec:analysis_6}
\uline{Question}: \textit{Are there identifiable lexical patterns in samples that are especially successfully classified by the adversarially trained model for emotion classification?}\\
\uline{Hypothesis}: \textit{Certain properties of input features correlate with the increase in probability of successful classification in adversarially trained emotion recognition models.}\\
Our results in questions 4 and 5 of section~\ref{sec:analysis_4} 
demonstrate that decorrelating the representations from modulations due to confounding factors can positively affect the classification performance of our trained emotion recognition models when applied to datasets whose properties differ from the data on which the models were trained.

In this section, we aim to understand what properties of input features in a given sample
correlate with the probability of successful classification in trained emotion recognition models due to decorrelation of such modulations. 
Understanding the relationship between the properties of the input features and the likelihood of success 
can help us identify data points that are more likely to be correctly classified using adversarially trained models. This can help us identify samples in an unseen dataset for which we can trust the classification label obtained from the adversarial model as compared to the normal classification model.
We analyze this relationship using word tokens, which are low-dimensional and human-interpretable.  
\subsubsection{\textbf{Adjusted Probability of Success}}
We study the correlation between the lexical patterns of data samples and the probability that those samples are correctly classified.  We focus on improvements in classifaction, moving from the normal model to the adversarial model.  This allows us to focus on improvement and mitigates the challenge that certain samples may just be particularly easy or hard to classify. 
We define probability of success for a sample as the $P_{A, s}(Success)$ where A can either be a normal (\textit{normal}) classification model or an adversarial classification model (\textit{adv}) and $s$ refers to the index of a particular sample.  We calculate $P_A(Success)$ as the ratio of the number of times a model correctly classifies a given sample to the total number of fifteen runs. 
If a sample is correctly classified across all runs by adversarial model, the $P_{adv}(Success)$ for that sample would be $1$. But we want to concentrate on the gain in performance of using adversarial over normal classification. It might be the case that this sample is correctly classified across all runs by normal classification model as well, the $P_{normal}(Success)$ for that sample would be $1$. In this case, we do not see any betterment as a result of using adversarial training paradigm.
To mitigate the above limitation, we define adjusted probability of success in the following manner:
We define adjusted probability of success (APS) for sample $s$ as: $P_{adv, s}(Success)-P_{norma, sl}(Success)$.  When the APS is greater than 0, the sample is more accurately classified using the adversarial model.  When the APS is less than zero, it is more accurately classifed using the normal model.

\subsubsection{\textbf{Features}}
Our goal is to correlate APS with interpretable lexical features. We 
use the Linguistic Inquiry and Word Count (LIWC)~\cite{pennebaker2001linguistic} tool. 
LIWC assigns a predefined category to a word based on its association with social, affective and cognitive process.
These categories have been shown to be highly predictive of both emotion~\cite{kahn2007measuring}, spontatenity~\cite{clark2002using} and stress~\cite{wang2016twitter}.

We form a twelve length feature vector for each utterance by counting the number of words that fall into each of the nine LIWC categories (adverb, pronoun, social process, negation, positive and negative emotion, insight, tentative and certainty).  We normalize the feature vector by how many words in the utterance.
We augment this feature vector to include: (1) fillers (e.g., ``uhh''),  hesitation (e.g., ``like''), and discourse markers (e.g., ``so'') and (2) content rate, defined as the number of words per unit length of time. The final feature vector comprises of all the above mentioned categories.
\subsubsection{\textbf{Discussion}}
\textbf{Decorrelating Stress.} We report the Pearson correlation coefficient and the resulting Benjamini-Hochberg adjusted~\cite{benjamini1995controlling} $p$-values that we obtain between each feature in the vector, and the APS for each sample.  We perform this assessment for both the activation and valence normal and adversarial lexical models.  We focus on the cross-dataset case in which the models were trained on MuSE and tested on IEMOCAP (in Table~\ref{exp-linguistic-correlation-stress}). A large positive correlation between a category and APS implies that samples with larger numbers of words in a given category are likely to be classified correctly more often given the adversarial model versus the normal model. 

\begin{table}[t]
\scriptsize
  \caption{\footnotesize Correlation between LIWC features and APS due to stress decorrelation, for activation and valence.
  $p$-values are Benjamini-Hochberg adjusted for multiple comparisons ($\alpha = 0.05$). p-value codes: `**'$<$0.01; `*'$<$0.05; `-'$<$0.1}
  \label{exp-linguistic-correlation-stress}
  \centering
  \begin{tabular}{lcccc}
    \toprule
    & \multicolumn{2}{c}{Act.} & \multicolumn{2}{c}{Val.}  \\
    \cmidrule(r){2-3} \cmidrule(r){4-5}
         & r     &  p  & r     &  p\\
    \midrule
    \textbf{LIWC} \\
    \midrule
    Adverb                      & \textbf{0.217}   & ** & \textbf{0.177}  &  *  \\
    Pronoun                     & \textbf{0.165}   & -  & 0.082 & * \\
    Social Process (social)     & 0.084  & -  &   0.001    & -      \\
    Negations (negate)          & -0.018   & -  & 0.005  & -\\
    Positive emotion (posemo)   & \textbf{0.154}    & *  &  0.093  & - \\
    Negative emotion (negemo)   & 0.086   & -  & \textbf{0.143} & * \\
    Insight                     & -0.021   & -  & -0.012   &   -   \\
    Tentative (tentat)          & 0.074   & -  &    0.101    &  -    \\
    Certainty (certain)         & \textbf{0.138} & *  &  0.116     &  -\\
    \midrule
    \textbf{Hesitation}          \\
    \midrule
    Fillers      &  \textbf{0.154} & *  & \textbf{0.182} & * \\
    Discourse marker &  \textbf{0.141}     & *  & 0.111 & - \\
    Content Rate    &   \textbf{0.196}    & ** & \textbf{0.178} & ** \\
    \bottomrule
  \end{tabular}
  \vspace{-1em}
\end{table}

\begin{itemize}
\item Activation recognition: APS is significantly positively correlated with the presence of words that relate to: Adverb (0.217), pronoun (0.165), positive emotion (0.154), certainity (0.138), fillers (0.154), discourse markers (0.141), and content rate (0.196).
\item Valence recognition: APS is significantly positively correlated with the presence of words that relate to: Adverb (0.177), negative emotion (0.143), fillers (0.182), content rate (0.178).

\end{itemize}
This finding is consistent with previous research~\cite{mehl2017natural}, where the authors have shown that there is often an  increase in usage of function words and intensifiers in stressed conditions. So, for example a sentence "I am really really sad about losing my pen" would have more likelihood of being correctly classified by the adversarial model compared to the normal emotion classification model. Hence, we can hypothesise that an increase in the likelihood of correct classification of samples containing these intensifiers occurs due to reduced weightage of adverbs in adversarial training paradigm.

There are fewer significant categories for valence than for activation. This is consistent with the results in Table~\ref{exp-tri-other} for the lexical modality.
Although we see a significant correlation between filler words and APS for activation classification~\cite{Duvall2014ExploringFW}, we do not observe the same for discourse markers and presence of social process words.  
The absence of significance in these cases implies that though these values are markers of stress, the normal classifier is still able to learn reliable representations invariant of stress for predicting the correct target label, resulting in negligible impact on classification performance when decorrelating the representations.



\textbf{Decorrelating Spontatenity.} We do a similar analysis for analyzing lexical properties of samples that were aided by decorrelating spontaneity. We report the Pearson correlation coefficient and the resulting Benjamini-Hochberg adjusted~\cite{benjamini1995controlling} $p$-values that we obtain from the LIWC features and APS for both emotion axes lexical-based classification models (trained on IEMOCAP; tested on MuSE) in Table~\ref{exp-linguistic-correlation-spont}.

\begin{table}[t]
\scriptsize
  \caption{\footnotesize Correlation between LIWC features and APS due to spontaenity decorrelation, for activation and valence.
  $p$-values are Benjamini-Hochberg adjusted for multiple comparisons ($\alpha = 0.05$). p-value codes: `**'$<$0.01; `*'$<$0.05; `-'$<$0.1}
  \label{exp-linguistic-correlation-spont}
  \centering
  \begin{tabular}{lcccc}
    \toprule
    & \multicolumn{2}{c}{Act.} & \multicolumn{2}{c}{Val.}  \\
    \cmidrule(r){2-3} \cmidrule(r){4-5}
         & r     &  p  & r     &  p\\
    \midrule
    \textbf{LIWC} \\
    \midrule
    Adverb                      & 0.121   & ** & 0.088  &  -  \\
    Pronoun                     & \textbf{0.138}   & -  & 0.016 & * \\
    Social Process (social)     & 0.132  & -  &   \textbf{0.166}     & *      \\
    Negations (negate)          & -0.003   & -  & -0.011  & -\\
    Positive emotion (posemo)   & 0.122   & -  &  \textbf{0.161}  & * \\
    Negative emotion (negemo)   & \textbf{0.137}   & *  & \textbf{0.148} & * \\
    Insight                     & 0.017   & -  & 0.099   &   -   \\
    Tentative (tentat)          & \textbf{0.155}   & *  &    0.112    &  -    \\
    Certainty (certain)         & \textbf{0.191}  & *  &  \textbf{0.172}     &  *\\
    \midrule
    \textbf{Hesitation}          \\
    \midrule
    Fillers      &  \textbf{0.221} & **  & 0.119 & * \\
    Discourse marker &  \textbf{0.189}  & *  & 0.122 & - \\
    Content Rate                &   \textbf{0.165}   & * & \textbf{0.144} & - \\
    \bottomrule
  \end{tabular}
  \vspace{-12pt}
\end{table}

\begin{itemize}

\item Activation recognition: APS is significantly positively correlated with the presence of words that relate to: Pronoun (0.138), negative emotion (0.137), tentativeness (0.155), certainty (0.191), fillers (0.221), discourse markers (0.189) and content rate (0.165).

\item Valence recognition: APS is significantly positively correlated with the presence of words that relate to: Social Process (0.166), positive (0.161) and negative (0.148) emotion, certainty (0.172), and content rate (0.144).
\end{itemize}

The results suggest that there are identifiable linguistic properties of samples whose likelihood of correct classification benefits from the model trained adversarially to decorrelate spontaneity and emotion representation as compared to normal classification model. This is especially true for the use of words in the certainty category for both emotion dimensions and all hesitation categories for activation. Spontaneous speech has been shown to have more of these words in~\cite{clark2002using}. 
Scripted content has been shown to have more exaggerated displays of emotion through words and facial expressions~\cite{jurgens2015effect}. Controlling for the weights assigned to words in positive and negative emotion categories using the adversarial model, leads to better classification of samples that are comprised of these word tokens.


\section{Conclusions}
This work focused on the interplay between stress and emotion in the context of automatic emotion recognition.
We first showed that the presence of stress affects the performance of emotion recognition models.
We then observed that these effects vary depending on modality (acoustic or lexical) and task (activation or valence classification). We then showed how decorrelating stress modulations from emotion representations aids the generalizability of the model. Next, we showed how a similar method could be used to control for variations due to spontaneity; facilitating the generalizability of the model.
Finally, we identified human interpretable lexical markers that correlate with successful generalization of the model; especially concentrating on samples that are aided by decorrelation of stress and emotion representation.

Our results suggest that an extraneous psychological factor, such as stress, can significantly impact the performance of emotion recognition systems both within and across datasets. As a result, extraneous psychological factors should be accounted for when collecting data for training emotion recognition systems, especially when being used to predict labels of data that may or may not be modulated by those same factors. We then show how proactive decorrelation of this confounder can improve generalization of the model to other dataset at time of deployment. 

For future work, we will consider the trade-off between cross-dataset generalizability and within-dataset performance
to make better informed
decisions about which invariances to enforce. We will look into how other factors such as trust levels, social setting and conversation topic) influences emotion expression, and how we can make models that are adaptable to such varying scenarios.
\section{Acknowledgements}
This material is based in part upon work supported by the Toyota Research Institute (``TRI''), the IBM PhD Fellowship Award, and by the National Science Foundation (NSF CAREER 1651740). 
Any opinions, findings, and conclusions or recommendations expressed in this material are those of the authors and do not necessarily reflect the views of the NSF, IBM, TRI, or any other Toyota entity.

\vspace{1em}

\bibliographystyle{ACM-Reference-Format}
\bibliography{acmart}


\begin{thebibliography}{46}


\ifx \showCODEN    \undefined \def \showCODEN     #1{\unskip}     \fi
\ifx \showDOI      \undefined \def \showDOI       #1{#1}\fi
\ifx \showISBNx    \undefined \def \showISBNx     #1{\unskip}     \fi
\ifx \showISBNxiii \undefined \def \showISBNxiii  #1{\unskip}     \fi
\ifx \showISSN     \undefined \def \showISSN      #1{\unskip}     \fi
\ifx \showLCCN     \undefined \def \showLCCN      #1{\unskip}     \fi
\ifx \shownote     \undefined \def \shownote      #1{#1}          \fi
\ifx \showarticletitle \undefined \def \showarticletitle #1{#1}   \fi
\ifx \showURL      \undefined \def \showURL       {\relax}        \fi
\providecommand\bibfield[2]{#2}
\providecommand\bibinfo[2]{#2}
\providecommand\natexlab[1]{#1}
\providecommand\showeprint[2][]{arXiv:#2}

\bibitem[\protect\citeauthoryear{Abdelwahab and Busso}{Abdelwahab and
  Busso}{2018}]%
        {abdelwahab2018domain}
\bibfield{author}{\bibinfo{person}{Mohammed Abdelwahab} {and}
  \bibinfo{person}{Carlos Busso}.} \bibinfo{year}{2018}\natexlab{}.
\newblock \showarticletitle{Domain adversarial for acoustic emotion
  recognition}.
\newblock \bibinfo{journal}{\emph{IEEE/ACM Transactions on Audio, Speech, and
  Language Processing}} \bibinfo{volume}{26}, \bibinfo{number}{12}
  (\bibinfo{year}{2018}), \bibinfo{pages}{2423--2435}.
\newblock


\bibitem[\protect\citeauthoryear{Aron, Melinat, Aron, Vallone, and Bator}{Aron
  et~al\mbox{.}}{1997}]%
        {aron1997experimental}
\bibfield{author}{\bibinfo{person}{Arthur Aron}, \bibinfo{person}{Edward
  Melinat}, \bibinfo{person}{Elaine~N Aron}, \bibinfo{person}{Robert~Darrin
  Vallone}, {and} \bibinfo{person}{Renee~J Bator}.}
  \bibinfo{year}{1997}\natexlab{}.
\newblock \showarticletitle{The experimental generation of interpersonal
  closeness: A procedure and some preliminary findings}.
\newblock \bibinfo{journal}{\emph{Personality and Social Psychology Bulletin}}
  \bibinfo{volume}{23}, \bibinfo{number}{4} (\bibinfo{year}{1997}),
  \bibinfo{pages}{363--377}.
\newblock


\bibitem[\protect\citeauthoryear{Ben-David, Blitzer, Crammer, Kulesza, Pereira,
  and Vaughan}{Ben-David et~al\mbox{.}}{2010}]%
        {ben2010theory}
\bibfield{author}{\bibinfo{person}{Shai Ben-David}, \bibinfo{person}{John
  Blitzer}, \bibinfo{person}{Koby Crammer}, \bibinfo{person}{Alex Kulesza},
  \bibinfo{person}{Fernando Pereira}, {and} \bibinfo{person}{Jennifer~Wortman
  Vaughan}.} \bibinfo{year}{2010}\natexlab{}.
\newblock \showarticletitle{A theory of learning from different domains}.
\newblock \bibinfo{journal}{\emph{Machine learning}} \bibinfo{volume}{79},
  \bibinfo{number}{1-2} (\bibinfo{year}{2010}), \bibinfo{pages}{151--175}.
\newblock


\bibitem[\protect\citeauthoryear{Benjamini and Hochberg}{Benjamini and
  Hochberg}{1995}]%
        {benjamini1995controlling}
\bibfield{author}{\bibinfo{person}{Yoav Benjamini} {and} \bibinfo{person}{Yosef
  Hochberg}.} \bibinfo{year}{1995}\natexlab{}.
\newblock \showarticletitle{Controlling the false discovery rate: a practical
  and powerful approach to multiple testing}.
\newblock \bibinfo{journal}{\emph{Journal of the Royal statistical society:
  series B (Methodological)}} \bibinfo{volume}{57}, \bibinfo{number}{1}
  (\bibinfo{year}{1995}), \bibinfo{pages}{289--300}.
\newblock


\bibitem[\protect\citeauthoryear{Buchanan, Laures-Gore, and Duff}{Buchanan
  et~al\mbox{.}}{2014}]%
        {buchanan2014acute}
\bibfield{author}{\bibinfo{person}{Tony~W Buchanan},
  \bibinfo{person}{Jacqueline~S Laures-Gore}, {and} \bibinfo{person}{Melissa~C
  Duff}.} \bibinfo{year}{2014}\natexlab{}.
\newblock \showarticletitle{Acute stress reduces speech fluency}.
\newblock \bibinfo{journal}{\emph{Biological psychology}}  \bibinfo{volume}{97}
  (\bibinfo{year}{2014}), \bibinfo{pages}{60--66}.
\newblock


\bibitem[\protect\citeauthoryear{Busso, Bulut, Lee, Kazemzadeh, Mower, Kim,
  Chang, Lee, and Narayanan}{Busso et~al\mbox{.}}{2008}]%
        {busso2008iemocap}
\bibfield{author}{\bibinfo{person}{Carlos Busso}, \bibinfo{person}{Murtaza
  Bulut}, \bibinfo{person}{Chi-Chun Lee}, \bibinfo{person}{Abe Kazemzadeh},
  \bibinfo{person}{Emily Mower}, \bibinfo{person}{Samuel Kim},
  \bibinfo{person}{Jeannette~N Chang}, \bibinfo{person}{Sungbok Lee}, {and}
  \bibinfo{person}{Shrikanth~S Narayanan}.} \bibinfo{year}{2008}\natexlab{}.
\newblock \showarticletitle{IEMOCAP: Interactive emotional dyadic motion
  capture database}.
\newblock \bibinfo{journal}{\emph{Language resources and evaluation}}
  \bibinfo{volume}{42}, \bibinfo{number}{4} (\bibinfo{year}{2008}),
  \bibinfo{pages}{335}.
\newblock


\bibitem[\protect\citeauthoryear{Busso, Parthasarathy, Burmania, AbdelWahab,
  Sadoughi, and Provost}{Busso et~al\mbox{.}}{2017}]%
        {busso2017msp}
\bibfield{author}{\bibinfo{person}{Carlos Busso}, \bibinfo{person}{Srinivas
  Parthasarathy}, \bibinfo{person}{Alec Burmania}, \bibinfo{person}{Mohammed
  AbdelWahab}, \bibinfo{person}{Najmeh Sadoughi}, {and}
  \bibinfo{person}{Emily~Mower Provost}.} \bibinfo{year}{2017}\natexlab{}.
\newblock \showarticletitle{MSP-IMPROV: An acted corpus of dyadic interactions
  to study emotion perception}.
\newblock \bibinfo{journal}{\emph{IEEE Transactions on Affective Computing}}
  \bibinfo{volume}{8}, \bibinfo{number}{1} (\bibinfo{year}{2017}),
  \bibinfo{pages}{67--80}.
\newblock


\bibitem[\protect\citeauthoryear{Chollet}{Chollet}{2015}]%
        {keras2015}
\bibfield{author}{\bibinfo{person}{François Chollet}.}
  \bibinfo{year}{2015}\natexlab{}.
\newblock \bibinfo{title}{keras}.
\newblock \bibinfo{howpublished}{\url{https://github.com/fchollet/keras}}.
\newblock


\bibitem[\protect\citeauthoryear{Clark and Tree}{Clark and Tree}{2002}]%
        {clark2002using}
\bibfield{author}{\bibinfo{person}{Herbert~H Clark} {and} \bibinfo{person}{Jean
  E~Fox Tree}.} \bibinfo{year}{2002}\natexlab{}.
\newblock \showarticletitle{Using uh and um in spontaneous speaking}.
\newblock \bibinfo{journal}{\emph{Cognition}} \bibinfo{volume}{84},
  \bibinfo{number}{1} (\bibinfo{year}{2002}), \bibinfo{pages}{73--111}.
\newblock


\bibitem[\protect\citeauthoryear{Cohen, Kamarck, and Mermelstein}{Cohen
  et~al\mbox{.}}{1983}]%
        {cohen1983global}
\bibfield{author}{\bibinfo{person}{Sheldon Cohen}, \bibinfo{person}{Tom
  Kamarck}, {and} \bibinfo{person}{Robin Mermelstein}.}
  \bibinfo{year}{1983}\natexlab{}.
\newblock \showarticletitle{A global measure of perceived stress}.
\newblock \bibinfo{journal}{\emph{Journal of health and social behavior}}
  (\bibinfo{year}{1983}), \bibinfo{pages}{385--396}.
\newblock


\bibitem[\protect\citeauthoryear{DeDora, Carlson, and Mujica-Parodi}{DeDora
  et~al\mbox{.}}{2011}]%
        {dedora2011acute}
\bibfield{author}{\bibinfo{person}{Daniel~J DeDora}, \bibinfo{person}{Joshua~M
  Carlson}, {and} \bibinfo{person}{Lilianne~R Mujica-Parodi}.}
  \bibinfo{year}{2011}\natexlab{}.
\newblock \showarticletitle{Acute stress eliminates female advantage in
  detection of ambiguous negative affect}.
\newblock \bibinfo{journal}{\emph{Evolutionary Psychology}}
  \bibinfo{volume}{9}, \bibinfo{number}{4} (\bibinfo{year}{2011}),
  \bibinfo{pages}{147470491100900406}.
\newblock


\bibitem[\protect\citeauthoryear{Duvall, Robbins, Graham, and Divett}{Duvall
  et~al\mbox{.}}{2014}]%
        {Duvall2014ExploringFW}
\bibfield{author}{\bibinfo{person}{Emily~D. Duvall},
  \bibinfo{person}{Alan~Stuart Robbins}, \bibinfo{person}{Thomas~R Graham},
  {and} \bibinfo{person}{Scott Divett}.} \bibinfo{year}{2014}\natexlab{}.
\newblock \showarticletitle{Exploring Filler Words and Their Impact}.
\newblock


\bibitem[\protect\citeauthoryear{Elazar and Goldberg}{Elazar and
  Goldberg}{2018}]%
        {elazar2018adversarial}
\bibfield{author}{\bibinfo{person}{Yanai Elazar} {and} \bibinfo{person}{Yoav
  Goldberg}.} \bibinfo{year}{2018}\natexlab{}.
\newblock \showarticletitle{Adversarial Removal of Demographic Attributes from
  Text Data}.
\newblock \bibinfo{journal}{\emph{arXiv preprint arXiv:1808.06640}}
  (\bibinfo{year}{2018}).
\newblock


\bibitem[\protect\citeauthoryear{Freeman, Hart, Kimbrell, and Ross}{Freeman
  et~al\mbox{.}}{2009}]%
        {freeman2009comprehension}
\bibfield{author}{\bibinfo{person}{Thomas~W Freeman}, \bibinfo{person}{John
  Hart}, \bibinfo{person}{Tim Kimbrell}, {and} \bibinfo{person}{Elliott~D
  Ross}.} \bibinfo{year}{2009}\natexlab{}.
\newblock \showarticletitle{Comprehension of affective prosody in veterans with
  chronic posttraumatic stress disorder}.
\newblock \bibinfo{journal}{\emph{The Journal of neuropsychiatry and clinical
  neurosciences}} \bibinfo{volume}{21}, \bibinfo{number}{1}
  (\bibinfo{year}{2009}), \bibinfo{pages}{52--58}.
\newblock


\bibitem[\protect\citeauthoryear{Ganin and Lempitsky}{Ganin and
  Lempitsky}{2014}]%
        {ganin2014unsupervised}
\bibfield{author}{\bibinfo{person}{Yaroslav Ganin} {and}
  \bibinfo{person}{Victor Lempitsky}.} \bibinfo{year}{2014}\natexlab{}.
\newblock \showarticletitle{Unsupervised domain adaptation by backpropagation}.
\newblock \bibinfo{journal}{\emph{arXiv preprint arXiv:1409.7495}}
  (\bibinfo{year}{2014}).
\newblock


\bibitem[\protect\citeauthoryear{Ganin, Ustinova, Ajakan, Germain, Larochelle,
  Laviolette, Marchand, and Lempitsky}{Ganin et~al\mbox{.}}{2016}]%
        {ganin2016domain}
\bibfield{author}{\bibinfo{person}{Yaroslav Ganin}, \bibinfo{person}{Evgeniya
  Ustinova}, \bibinfo{person}{Hana Ajakan}, \bibinfo{person}{Pascal Germain},
  \bibinfo{person}{Hugo Larochelle}, \bibinfo{person}{Fran{\c{c}}ois
  Laviolette}, \bibinfo{person}{Mario Marchand}, {and} \bibinfo{person}{Victor
  Lempitsky}.} \bibinfo{year}{2016}\natexlab{}.
\newblock \showarticletitle{Domain-adversarial training of neural networks}.
\newblock \bibinfo{journal}{\emph{The Journal of Machine Learning Research}}
  \bibinfo{volume}{17}, \bibinfo{number}{1} (\bibinfo{year}{2016}),
  \bibinfo{pages}{2096--2030}.
\newblock


\bibitem[\protect\citeauthoryear{Gideon, McInnis, and Mower~Provost}{Gideon
  et~al\mbox{.}}{2019}]%
        {gideon2019barking}
\bibfield{author}{\bibinfo{person}{John Gideon}, \bibinfo{person}{Melvin~G
  McInnis}, {and} \bibinfo{person}{Emily Mower~Provost}.}
  \bibinfo{year}{2019}\natexlab{}.
\newblock \showarticletitle{Barking up the Right Tree: Improving Cross-Corpus
  Speech Emotion Recognition with Adversarial Discriminative Domain
  Generalization (ADDoG)}.
\newblock \bibinfo{journal}{\emph{arXiv preprint arXiv:1903.12094}}
  (\bibinfo{year}{2019}).
\newblock


\bibitem[\protect\citeauthoryear{Hu, Peng, Yang, Hospedales, and Verbeek}{Hu
  et~al\mbox{.}}{2017}]%
        {hu2017frankenstein}
\bibfield{author}{\bibinfo{person}{Guosheng Hu}, \bibinfo{person}{Xiaojiang
  Peng}, \bibinfo{person}{Yongxin Yang}, \bibinfo{person}{Timothy~M
  Hospedales}, {and} \bibinfo{person}{Jakob Verbeek}.}
  \bibinfo{year}{2017}\natexlab{}.
\newblock \showarticletitle{Frankenstein: Learning deep face representations
  using small data}.
\newblock \bibinfo{journal}{\emph{IEEE Transactions on Image Processing}}
  \bibinfo{volume}{27}, \bibinfo{number}{1} (\bibinfo{year}{2017}),
  \bibinfo{pages}{293--303}.
\newblock


\bibitem[\protect\citeauthoryear{Jaiswal, Aldeneh, Bara, Luo, Burzo, Mihalcea,
  and Mower~Provost}{Jaiswal et~al\mbox{.}}{2019}]%
        {jaiswal2019muse}
\bibfield{author}{\bibinfo{person}{Mimansa Jaiswal}, \bibinfo{person}{Zakaria
  Aldeneh}, \bibinfo{person}{Cristian-Paul Bara}, \bibinfo{person}{Yuanhang
  Luo}, \bibinfo{person}{Mihai Burzo}, \bibinfo{person}{Rada Mihalcea}, {and}
  \bibinfo{person}{Emily Mower~Provost}.} \bibinfo{year}{2019}\natexlab{}.
\newblock \showarticletitle{MuSE-ing on the impact of utterance ordering on
  crowdsourced emotion annotations}. In \bibinfo{booktitle}{\emph{2019 IEEE
  International Conference on Acoustics, Speech and Signal Processing
  (ICASSP)}}. IEEE.
\newblock


\bibitem[\protect\citeauthoryear{J{\"u}rgens, Grass, Drolet, and
  Fischer}{J{\"u}rgens et~al\mbox{.}}{2015}]%
        {jurgens2015effect}
\bibfield{author}{\bibinfo{person}{Rebecca J{\"u}rgens},
  \bibinfo{person}{Annika Grass}, \bibinfo{person}{Matthis Drolet}, {and}
  \bibinfo{person}{Julia Fischer}.} \bibinfo{year}{2015}\natexlab{}.
\newblock \showarticletitle{Effect of acting experience on emotion expression
  and recognition in voice: Non-actors provide better stimuli than expected}.
\newblock \bibinfo{journal}{\emph{Journal of nonverbal behavior}}
  \bibinfo{volume}{39}, \bibinfo{number}{3} (\bibinfo{year}{2015}),
  \bibinfo{pages}{195--214}.
\newblock


\bibitem[\protect\citeauthoryear{Kahn, Tobin, Massey, and Anderson}{Kahn
  et~al\mbox{.}}{2007}]%
        {kahn2007measuring}
\bibfield{author}{\bibinfo{person}{Jeffrey~H Kahn}, \bibinfo{person}{Renee~M
  Tobin}, \bibinfo{person}{Audra~E Massey}, {and} \bibinfo{person}{Jennifer~A
  Anderson}.} \bibinfo{year}{2007}\natexlab{}.
\newblock \showarticletitle{Measuring emotional expression with the Linguistic
  Inquiry and Word Count}.
\newblock \bibinfo{journal}{\emph{The American journal of psychology}}
  (\bibinfo{year}{2007}), \bibinfo{pages}{263--286}.
\newblock


\bibitem[\protect\citeauthoryear{Khorram, Aldeneh, Dimitriadis, McInnis, and
  Provost}{Khorram et~al\mbox{.}}{2017}]%
        {khorram2017capturing}
\bibfield{author}{\bibinfo{person}{Soheil Khorram}, \bibinfo{person}{Zakaria
  Aldeneh}, \bibinfo{person}{Dimitrios Dimitriadis}, \bibinfo{person}{Melvin
  McInnis}, {and} \bibinfo{person}{Emily~Mower Provost}.}
  \bibinfo{year}{2017}\natexlab{}.
\newblock \showarticletitle{Capturing long-term temporal dependencies with
  convolutional networks for continuous emotion recognition}.
\newblock \bibinfo{journal}{\emph{arXiv preprint arXiv:1708.07050}}
  (\bibinfo{year}{2017}).
\newblock


\bibitem[\protect\citeauthoryear{Khorram, Jaiswal, Gideon, McInnis, and
  Provost}{Khorram et~al\mbox{.}}{2018}]%
        {khorram2018priori}
\bibfield{author}{\bibinfo{person}{Soheil Khorram}, \bibinfo{person}{Mimansa
  Jaiswal}, \bibinfo{person}{John Gideon}, \bibinfo{person}{Melvin McInnis},
  {and} \bibinfo{person}{Emily~Mower Provost}.}
  \bibinfo{year}{2018}\natexlab{}.
\newblock \showarticletitle{The PRIORI Emotion Dataset: Linking Mood to Emotion
  Detected In-the-Wild}.
\newblock \bibinfo{journal}{\emph{arXiv preprint arXiv:1806.10658}}
  (\bibinfo{year}{2018}).
\newblock


\bibitem[\protect\citeauthoryear{Kim}{Kim}{2014}]%
        {kim2014convolutional}
\bibfield{author}{\bibinfo{person}{Yoon Kim}.} \bibinfo{year}{2014}\natexlab{}.
\newblock \showarticletitle{Convolutional neural networks for sentence
  classification}.
\newblock \bibinfo{journal}{\emph{arXiv preprint arXiv:1408.5882}}
  (\bibinfo{year}{2014}).
\newblock


\bibitem[\protect\citeauthoryear{Kol{\'a}{\v{r}}}{Kol{\'a}{\v{r}}}{2008}]%
        {kolavr2008automatic}
\bibfield{author}{\bibinfo{person}{J{\'a}chym Kol{\'a}{\v{r}}}.}
  \bibinfo{year}{2008}\natexlab{}.
\newblock \emph{\bibinfo{title}{Automatic Segmentation of Speech into
  Sentence-like Units}}.
\newblock \bibinfo{thesistype}{Ph.D. Dissertation}. \bibinfo{school}{University
  of West Bohemia in Pilsen}.
\newblock


\bibitem[\protect\citeauthoryear{Krishna, Lu, Gimpel, and Livescu}{Krishna
  et~al\mbox{.}}{2018}]%
        {krishna2018study}
\bibfield{author}{\bibinfo{person}{Kalpesh Krishna}, \bibinfo{person}{Liang
  Lu}, \bibinfo{person}{Kevin Gimpel}, {and} \bibinfo{person}{Karen Livescu}.}
  \bibinfo{year}{2018}\natexlab{}.
\newblock \showarticletitle{A Study of All-Convolutional Encoders for
  Connectionist Temporal Classification}. In \bibinfo{booktitle}{\emph{2018
  IEEE International Conference on Acoustics, Speech and Signal Processing
  (ICASSP)}}. IEEE, \bibinfo{pages}{5814--5818}.
\newblock


\bibitem[\protect\citeauthoryear{Landeiro and Culotta}{Landeiro and
  Culotta}{2016}]%
        {landeiro2016robust}
\bibfield{author}{\bibinfo{person}{Virgile Landeiro} {and}
  \bibinfo{person}{Aron Culotta}.} \bibinfo{year}{2016}\natexlab{}.
\newblock \showarticletitle{Robust text classification in the presence of
  confounding bias}. In \bibinfo{booktitle}{\emph{Thirtieth AAAI Conference on
  Artificial Intelligence}}.
\newblock


\bibitem[\protect\citeauthoryear{Locatello, Bauer, Lucic, Gelly, Sch{\"o}lkopf,
  and Bachem}{Locatello et~al\mbox{.}}{2018}]%
        {locatello2018challenging}
\bibfield{author}{\bibinfo{person}{Francesco Locatello},
  \bibinfo{person}{Stefan Bauer}, \bibinfo{person}{Mario Lucic},
  \bibinfo{person}{Sylvain Gelly}, \bibinfo{person}{Bernhard Sch{\"o}lkopf},
  {and} \bibinfo{person}{Olivier Bachem}.} \bibinfo{year}{2018}\natexlab{}.
\newblock \showarticletitle{Challenging common assumptions in the unsupervised
  learning of disentangled representations}.
\newblock \bibinfo{journal}{\emph{arXiv preprint arXiv:1811.12359}}
  (\bibinfo{year}{2018}).
\newblock


\bibitem[\protect\citeauthoryear{Mangalam and Guha}{Mangalam and Guha}{2017}]%
        {mangalam2017learning}
\bibfield{author}{\bibinfo{person}{Karttikeya Mangalam} {and}
  \bibinfo{person}{Tanaya Guha}.} \bibinfo{year}{2017}\natexlab{}.
\newblock \showarticletitle{Learning Spontaneity to Improve Emotion Recognition
  in Speech}.
\newblock \bibinfo{journal}{\emph{arXiv preprint arXiv:1712.04753}}
  (\bibinfo{year}{2017}).
\newblock


\bibitem[\protect\citeauthoryear{McHardy, Adel, and Klinger}{McHardy
  et~al\mbox{.}}{2019a}]%
        {mchardy2019adversarial}
\bibfield{author}{\bibinfo{person}{Robert McHardy}, \bibinfo{person}{Heike
  Adel}, {and} \bibinfo{person}{Roman Klinger}.}
  \bibinfo{year}{2019}\natexlab{a}.
\newblock \showarticletitle{Adversarial Training for Satire Detection:
  Controlling for Confounding Variables}.
\newblock \bibinfo{journal}{\emph{arXiv preprint arXiv:1902.11145}}
  (\bibinfo{year}{2019}).
\newblock


\bibitem[\protect\citeauthoryear{McHardy, Adel, and Klinger}{McHardy
  et~al\mbox{.}}{2019b}]%
        {satire-adv}
\bibfield{author}{\bibinfo{person}{Robert McHardy}, \bibinfo{person}{Heike
  Adel}, {and} \bibinfo{person}{Roman Klinger}.}
  \bibinfo{year}{2019}\natexlab{b}.
\newblock \showarticletitle{Adversarial Training for Satire Detection:
  Controlling for Confounding Variables}.
\newblock \bibinfo{journal}{\emph{CoRR}}  \bibinfo{volume}{abs/1902.11145}
  (\bibinfo{year}{2019}).
\newblock
\showeprint[arxiv]{1902.11145}
\urldef\tempurl%
\url{http://arxiv.org/abs/1902.11145}
\showURL{%
\tempurl}


\bibitem[\protect\citeauthoryear{Mehl, Raison, Pace, Arevalo, and Cole}{Mehl
  et~al\mbox{.}}{2017}]%
        {mehl2017natural}
\bibfield{author}{\bibinfo{person}{Matthias~R Mehl}, \bibinfo{person}{Charles~L
  Raison}, \bibinfo{person}{Thaddeus~WW Pace}, \bibinfo{person}{Jesusa~MG
  Arevalo}, {and} \bibinfo{person}{Steve~W Cole}.}
  \bibinfo{year}{2017}\natexlab{}.
\newblock \showarticletitle{Natural language indicators of differential gene
  regulation in the human immune system}.
\newblock \bibinfo{journal}{\emph{Proceedings of the National Academy of
  Sciences}} \bibinfo{volume}{114}, \bibinfo{number}{47}
  (\bibinfo{year}{2017}), \bibinfo{pages}{12554--12559}.
\newblock


\bibitem[\protect\citeauthoryear{Meng, Li, Chen, Zhao, Mazalov, Gang, and
  Juang}{Meng et~al\mbox{.}}{2018}]%
        {meng2018speaker}
\bibfield{author}{\bibinfo{person}{Zhong Meng}, \bibinfo{person}{Jinyu Li},
  \bibinfo{person}{Zhuo Chen}, \bibinfo{person}{Yang Zhao},
  \bibinfo{person}{Vadim Mazalov}, \bibinfo{person}{Yifan Gang}, {and}
  \bibinfo{person}{Biing-Hwang Juang}.} \bibinfo{year}{2018}\natexlab{}.
\newblock \showarticletitle{Speaker-invariant training via adversarial
  learning}. In \bibinfo{booktitle}{\emph{2018 IEEE International Conference on
  Acoustics, Speech and Signal Processing (ICASSP)}}. IEEE,
  \bibinfo{pages}{5969--5973}.
\newblock


\bibitem[\protect\citeauthoryear{Mikolov, Sutskever, Chen, Corrado, and
  Dean}{Mikolov et~al\mbox{.}}{2013}]%
        {mikolov2013distributed}
\bibfield{author}{\bibinfo{person}{Tomas Mikolov}, \bibinfo{person}{Ilya
  Sutskever}, \bibinfo{person}{Kai Chen}, \bibinfo{person}{Greg~S Corrado},
  {and} \bibinfo{person}{Jeff Dean}.} \bibinfo{year}{2013}\natexlab{}.
\newblock \showarticletitle{Distributed representations of words and phrases
  and their compositionality}. In \bibinfo{booktitle}{\emph{Advances in neural
  information processing systems}}. \bibinfo{pages}{3111--3119}.
\newblock


\bibitem[\protect\citeauthoryear{Molchanov, Tyree, Karras, Aila, and
  Kautz}{Molchanov et~al\mbox{.}}{2016}]%
        {molchanov2016pruning}
\bibfield{author}{\bibinfo{person}{Pavlo Molchanov}, \bibinfo{person}{Stephen
  Tyree}, \bibinfo{person}{Tero Karras}, \bibinfo{person}{Timo Aila}, {and}
  \bibinfo{person}{Jan Kautz}.} \bibinfo{year}{2016}\natexlab{}.
\newblock \showarticletitle{Pruning convolutional neural networks for resource
  efficient transfer learning}.
\newblock \bibinfo{journal}{\emph{arXiv preprint arXiv:1611.06440}}
  \bibinfo{volume}{3} (\bibinfo{year}{2016}).
\newblock


\bibitem[\protect\citeauthoryear{Moore, Pfeiffer, Wei, Iyer, Charles,
  Gilad-Bachrach, Boyles, and Manavoglu}{Moore et~al\mbox{.}}{2018}]%
        {moore2018modeling}
\bibfield{author}{\bibinfo{person}{John Moore}, \bibinfo{person}{Joel
  Pfeiffer}, \bibinfo{person}{Kai Wei}, \bibinfo{person}{Rishabh Iyer},
  \bibinfo{person}{Denis Charles}, \bibinfo{person}{Ran Gilad-Bachrach},
  \bibinfo{person}{Levi Boyles}, {and} \bibinfo{person}{Eren Manavoglu}.}
  \bibinfo{year}{2018}\natexlab{}.
\newblock \showarticletitle{Modeling and Simultaneously Removing Bias via
  Adversarial Neural Networks}.
\newblock \bibinfo{journal}{\emph{arXiv preprint arXiv:1804.06909}}
  (\bibinfo{year}{2018}).
\newblock


\bibitem[\protect\citeauthoryear{Paulmann, Furnes, B{\o}kenes, and
  Cozzolino}{Paulmann et~al\mbox{.}}{2016}]%
        {paulmann2016psychological}
\bibfield{author}{\bibinfo{person}{Silke Paulmann}, \bibinfo{person}{Desire
  Furnes}, \bibinfo{person}{Anne~Ming B{\o}kenes}, {and}
  \bibinfo{person}{Philip~J Cozzolino}.} \bibinfo{year}{2016}\natexlab{}.
\newblock \showarticletitle{How psychological stress affects emotional
  prosody}.
\newblock \bibinfo{journal}{\emph{Plos one}} \bibinfo{volume}{11},
  \bibinfo{number}{11} (\bibinfo{year}{2016}), \bibinfo{pages}{e0165022}.
\newblock


\bibitem[\protect\citeauthoryear{Pennebaker, Francis, and Booth}{Pennebaker
  et~al\mbox{.}}{2001}]%
        {pennebaker2001linguistic}
\bibfield{author}{\bibinfo{person}{James~W Pennebaker},
  \bibinfo{person}{Martha~E Francis}, {and} \bibinfo{person}{Roger~J Booth}.}
  \bibinfo{year}{2001}\natexlab{}.
\newblock \showarticletitle{Linguistic inquiry and word count: LIWC 2001}.
\newblock \bibinfo{journal}{\emph{Mahway: Lawrence Erlbaum Associates}}
  \bibinfo{volume}{71}, \bibinfo{number}{2001} (\bibinfo{year}{2001}),
  \bibinfo{pages}{2001}.
\newblock


\bibitem[\protect\citeauthoryear{Rosenberg}{Rosenberg}{2012}]%
        {rosenberg2012classifying}
\bibfield{author}{\bibinfo{person}{Andrew Rosenberg}.}
  \bibinfo{year}{2012}\natexlab{}.
\newblock \showarticletitle{Classifying skewed data: Importance weighting to
  optimize average recall}. In \bibinfo{booktitle}{\emph{Thirteenth Annual
  Conference of the International Speech Communication Association}}.
\newblock


\bibitem[\protect\citeauthoryear{Shinohara}{Shinohara}{2016}]%
        {shinohara2016adversarial}
\bibfield{author}{\bibinfo{person}{Yusuke Shinohara}.}
  \bibinfo{year}{2016}\natexlab{}.
\newblock \showarticletitle{Adversarial Multi-Task Learning of Deep Neural
  Networks for Robust Speech Recognition.}. In
  \bibinfo{booktitle}{\emph{INTERSPEECH}}. San Francisco, CA, USA,
  \bibinfo{pages}{2369--2372}.
\newblock


\bibitem[\protect\citeauthoryear{Steeneken and Hansen}{Steeneken and
  Hansen}{1999}]%
        {steeneken1999speech}
\bibfield{author}{\bibinfo{person}{Herman~JM Steeneken} {and}
  \bibinfo{person}{John~HL Hansen}.} \bibinfo{year}{1999}\natexlab{}.
\newblock \showarticletitle{Speech under stress conditions: overview of the
  effect on speech production and on system performance}. In
  \bibinfo{booktitle}{\emph{1999 IEEE International Conference on Acoustics,
  Speech, and Signal Processing. Proceedings. ICASSP99 (Cat. No. 99CH36258)}},
  Vol.~\bibinfo{volume}{4}. IEEE, \bibinfo{pages}{2079--2082}.
\newblock


\bibitem[\protect\citeauthoryear{Subbaswamy, Schulam, and Saria}{Subbaswamy
  et~al\mbox{.}}{2018}]%
        {subbaswamy2018learning}
\bibfield{author}{\bibinfo{person}{Adarsh Subbaswamy}, \bibinfo{person}{Peter
  Schulam}, {and} \bibinfo{person}{Suchi Saria}.}
  \bibinfo{year}{2018}\natexlab{}.
\newblock \showarticletitle{Learning predictive models that transport}.
\newblock \bibinfo{journal}{\emph{arXiv preprint arXiv:1812.04597}}
  (\bibinfo{year}{2018}).
\newblock


\bibitem[\protect\citeauthoryear{Tieleman and Hinton}{Tieleman and
  Hinton}{2012}]%
        {rmsprop}
\bibfield{author}{\bibinfo{person}{T. Tieleman} {and} \bibinfo{person}{G.
  Hinton}.} \bibinfo{year}{2012}\natexlab{}.
\newblock \bibinfo{title}{{Lecture 6.5---RmsProp: Divide the gradient by a
  running average of its recent magnitude}}.
\newblock \bibinfo{howpublished}{COURSERA: Neural Networks for Machine
  Learning}.
\newblock


\bibitem[\protect\citeauthoryear{Tzeng, Hoffman, Saenko, and Darrell}{Tzeng
  et~al\mbox{.}}{2017}]%
        {tzeng2017adversarial}
\bibfield{author}{\bibinfo{person}{Eric Tzeng}, \bibinfo{person}{Judy Hoffman},
  \bibinfo{person}{Kate Saenko}, {and} \bibinfo{person}{Trevor Darrell}.}
  \bibinfo{year}{2017}\natexlab{}.
\newblock \showarticletitle{Adversarial discriminative domain adaptation}. In
  \bibinfo{booktitle}{\emph{Proceedings of the IEEE Conference on Computer
  Vision and Pattern Recognition}}. \bibinfo{pages}{7167--7176}.
\newblock


\bibitem[\protect\citeauthoryear{Wang, Hernandez, Newman, He, and Bian}{Wang
  et~al\mbox{.}}{2016}]%
        {wang2016twitter}
\bibfield{author}{\bibinfo{person}{Wei Wang}, \bibinfo{person}{Ivan Hernandez},
  \bibinfo{person}{Daniel~A Newman}, \bibinfo{person}{Jibo He}, {and}
  \bibinfo{person}{Jiang Bian}.} \bibinfo{year}{2016}\natexlab{}.
\newblock \showarticletitle{Twitter analysis: Studying US weekly trends in work
  stress and emotion}.
\newblock \bibinfo{journal}{\emph{Applied Psychology}} \bibinfo{volume}{65},
  \bibinfo{number}{2} (\bibinfo{year}{2016}), \bibinfo{pages}{355--378}.
\newblock


\bibitem[\protect\citeauthoryear{Zhang, Lemoine, and Mitchell}{Zhang
  et~al\mbox{.}}{2018}]%
        {zhang2018mitigating}
\bibfield{author}{\bibinfo{person}{Brian~Hu Zhang}, \bibinfo{person}{Blake
  Lemoine}, {and} \bibinfo{person}{Margaret Mitchell}.}
  \bibinfo{year}{2018}\natexlab{}.
\newblock \showarticletitle{Mitigating unwanted biases with adversarial
  learning}. In \bibinfo{booktitle}{\emph{Proceedings of the 2018 AAAI/ACM
  Conference on AI, Ethics, and Society}}. ACM, \bibinfo{pages}{335--340}.
\newblock


\end{thebibliography}

\end{document}